\pdfoutput=1

\documentclass[11pt]{article}

\usepackage[]{acl}

\usepackage{times}
\usepackage{latexsym}
\usepackage{graphicx}
\usepackage{multirow}
\usepackage{booktabs} %
\usepackage{subcaption}
\usepackage{arydshln}
\usepackage{color, soul}
\definecolor{bananayellow}{rgb}{1.0, 0.88, 0.21}
\definecolor{skyblue}{rgb}{0.65, 0.78, 0.91}

\definecolor{lightgreen}{RGB}{172, 255, 146}
\definecolor{indigo}{RGB}{166, 196, 255}
\definecolor{pink}{RGB}{255, 162, 137}
\definecolor{yellow}{RGB}{255, 243, 128}
\newcommand{\hlgreen}[1]{{\sethlcolor{lightgreen}\hl{#1}}}
\newcommand{\hlindigo}[1]{{\sethlcolor{indigo}\hl{#1}}}
\newcommand{\hlpink}[1]{{\sethlcolor{pink}\hl{#1}}}
\newcommand{\hlyellow}[1]{{\sethlcolor{yellow}\hl{#1}}}

\usepackage[T1]{fontenc}

\usepackage[utf8]{inputenc}

\usepackage{microtype}

\newcommand{\taskname}{Description Set Generation (DSG)\space}
\newcommand{\taskacronym}{DSG\space}
\newcommand{\taskacronymnospace}{DSG}
\newcommand{\code}{\texttt}
\newcommand{\setAname}{\textit{generate-stratify}}
\newcommand{\setBname}{\textit{extract-stratify}}
\newcommand{\setCname}{\textit{generate-naive}}
\title{ACCoRD: A Multi-Document Approach to \\ Generating Diverse Descriptions of Scientific Concepts}

\author{Sonia K. Murthy, Kyle Lo, Daniel King, Chandra Bhagavatula, Bailey Kuehl \\ {\bf Sophie Johnson, Jonathan Borchardt, Daniel S. Weld, Tom Hope, Doug Downey} \\
  Allen Institute for Artificial Intelligence \\
  \{\tt soniam,kylel,daniel,chandrab,baileyk, 
  \\ \tt sophiej,jonathanb,danw,tomh,dougd\}@allenai.org}

\begin{document}
\maketitle

\begin{abstract}

Systems that can automatically define unfamiliar terms hold the promise of improving the accessibility of scientific texts, especially for readers who may lack prerequisite background knowledge. However, current systems assume a single ``best'' description per concept, which fails to account for the many potentially useful ways a concept can be described. We present ACCoRD, an end-to-end system tackling the novel task of generating \textit{sets} of descriptions of scientific concepts. Our system takes advantage of the myriad ways a concept is mentioned across the scientific literature to produce distinct, diverse descriptions of target scientific concepts in terms of different reference concepts.  To support research on the task, we release an expert-annotated resource, the ACCoRD corpus, which includes 1,275 labeled contexts and 1,787 hand-authored concept descriptions. We conduct a user study demonstrating that (1) users prefer descriptions produced by our end-to-end system, and (2) users prefer multiple descriptions to a single ``best'' description.

\end{abstract}

\section{Introduction}

Readers of scientific papers often encounter unfamiliar concepts, which impedes their understanding \cite{Portenoy2021BurstingSF}. This is because papers assume {\em a priori} knowledge, and often lack definitions for the scientific terms that they use. While readers may turn to external encyclopedic resources like Wikipedia, these contain descriptions for only a small fraction of scientific concepts \cite{forecite}, which has motivated the development of systems that automatically extract or generate descriptions for scientific concepts.  Unfortunately, current systems only surface a single ``best'' result for all users, which is often extracted from a single input document ~\cite{scidef2,scidef3,scidef4,heddex}.  The one-best description may not be accessible for all readers, given varying background knowledge.

\begin{figure}[t!]
    \centering
    \includegraphics[width=.45\textwidth]{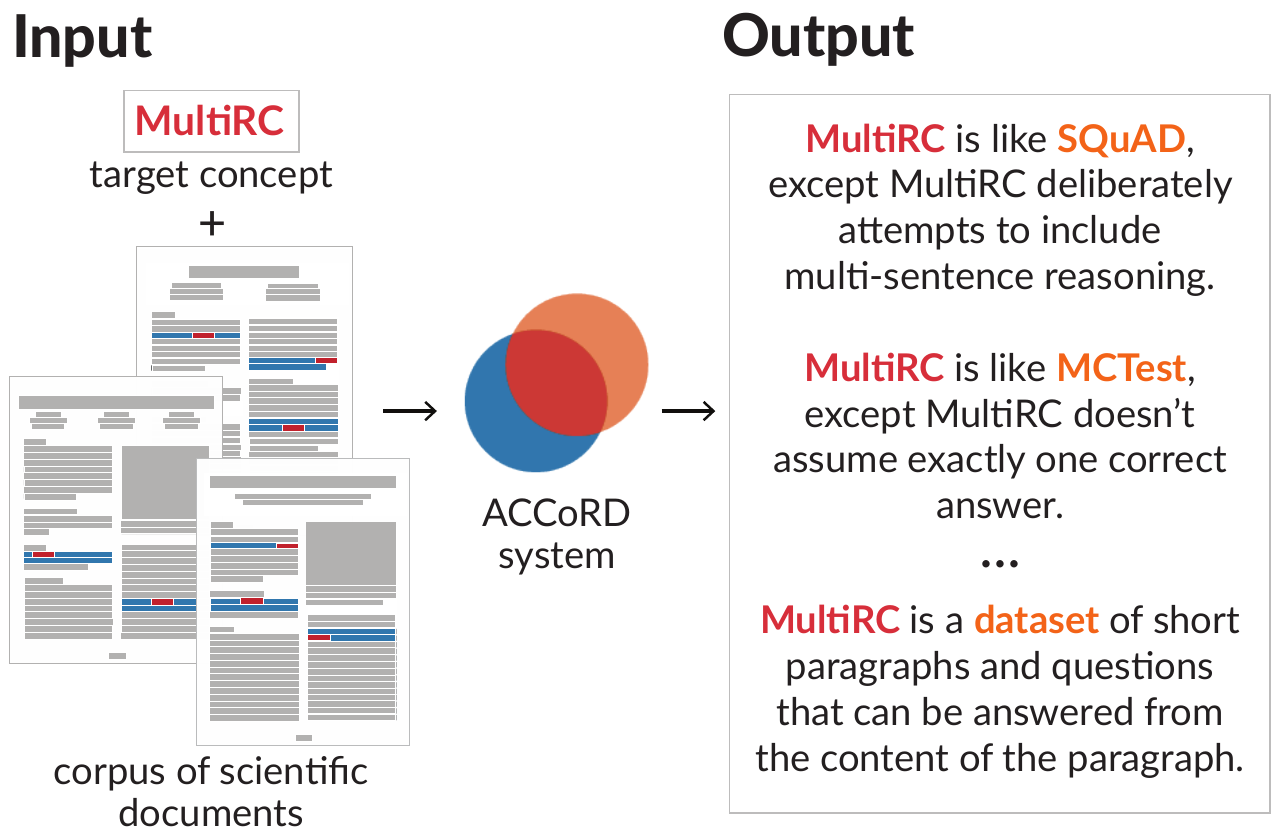}
    \caption{\textit{ACCoRD's approach to the Description Set Generation task}. Given a corpus of papers and a target concept to be described (red, e.g. MultiRC), our system produces a diverse \textit{set} of descriptions. 
    These are generated using mentions of the target concept in terms of myriad other reference concepts (orange) from extracted contexts (blue), resulting in a diverse set of descriptions.}
    \label{fig:input-output}
\end{figure}

\begin{figure*}[t!]
    \centering
    \includegraphics[width=\textwidth]{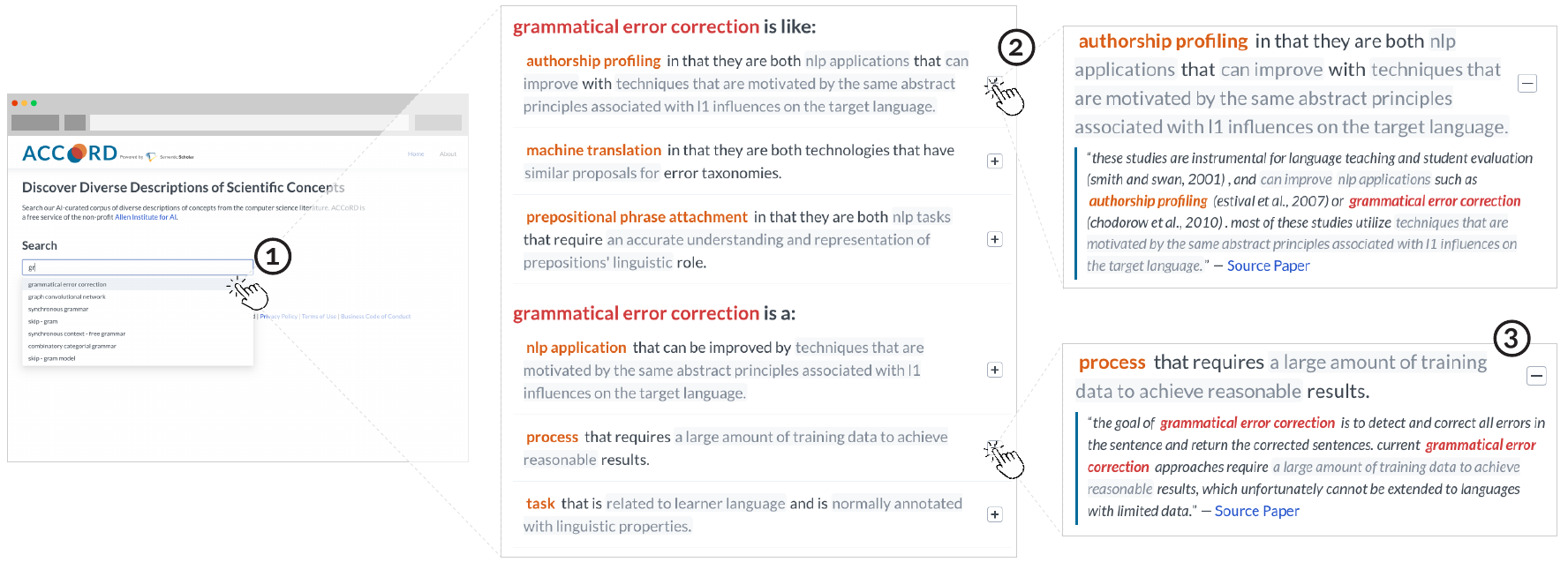}
    \caption{\textit{Demo screenshots.} (1) Users search for a target scientific concept from a pre-defined list and are shown cards for top reference concepts used to describe the target concept in terms of a particular relation. (2) Users click to expand the cards to see the extracted snippet (context) that produced the generated concept description, in addition to a link to the source paper. (3) Spans of text that are shared between the extracted context and generated description are highlighted to facilitate easy comparison.}
    \label{fig:system-screenshots}
\end{figure*}

Scientific concepts can be described in multiple distinct ways, and our approach is based on our hypothesis that a {\em set of descriptions} is more useful for users than a single description.
Humans learn new concepts by understanding how they relate to other, known concepts \cite{RumelhartOrtony1977,Spiro1980,nrc2018}, and providing multiple descriptions allows us to highlight multiple such relationships, contributing to a more complete understanding. 
Furthermore, providing multiple descriptions increases the number of potentially helpful connections between a new concept and concepts within the user's specific background knowledge (see Figure \ref{fig:input-output}), increasing accessibility.
This relational approach to human concept learning has been formalized through the lens of Analogical Transfer Theory \cite{Gentner1983StructureMappingAT, analogicalTheory2001, gentner2003language} and has long been employed as a tool in scientific discourse and education \cite{Treagust1992ScienceTU, Heywood2002ThePO}. 
Our work expands upon the notion of a description in the context of description generation systems to include analogy-like descriptions that are currently not captured by either scientific definition \cite{heddex} or relation extraction \cite{dygie} systems.

In this work, we present Automatic Comparison of Concepts with Relational Descriptions (ACCoRD) -- an end-to-end system that tackles the novel task of producing a set of distinct descriptions for a given target concept.\footnote{System demo, code, and data set available at
\href{http://github.com/allenai/ACCoRD}{github.com/allenai/ACCoRD}}
Given text from scientific papers, our system first {\textbf{extracts}} all sentences from across the corpus that contain a description of the concept in terms of any other concept. Then, conditioned on the extractions, ACCoRD {\textbf{generates}} succinct, self-contained descriptions of the concepts' relationship using GPT-3 \cite{brown2020language} in the few-shot setting.
The system finally {\textbf{selects}} a smaller, yet diverse subset of descriptions that captures the richness of a concept's usages by including multiple relation types and reference concepts.

Our contributions are:
\begin{enumerate}
    \item We introduce Description Set Generation (DSG), the novel task of generating multiple distinct descriptions of a single target concept. In support of this task, we release the ACCoRD corpus, an expert-annotated resource of 1,275 labeled contexts and 1,787 hand-authored concept descriptions.
    \item We present ACCoRD, an end-to-end system for DSG that outputs a diverse set of descriptions for concepts in the computer science domain. %
    \item We conduct a user study demonstrating that users prefer multiple descriptions over a single ``best'' description, and that they prefer our system's generated concept descriptions over those of an extractive baseline.
\end{enumerate}

\section{Description Set Generation}

\subsection{Task definition} We introduce the task of \taskname as follows: Given a large corpus of $N$ scientific documents, a target concept to be described and a desired output size $|S|$, the goal of \taskacronym is to output a set $S$ of succinct, self-contained, and distinct descriptions of the target concept. A schematic of this task is provided in Figure~\ref{fig:input-output}. Unlike prior work, which defines the task in terms of a single output description per scientific concept \cite{w00,heddex}, DSG proposes outputting a set of descriptions. One can view DSG as a generalization of the format used in prior work (i.e., single-description outputs are sets with $|S| = 1$).

\subsection{Approach}
DSG is an open-ended task, and many possible description sets could form valid output for a given concept. 
To facilitate the generation of descriptions that are useful and factual, in this work, we focus on descriptions that meet three criteria: (i) They are derived from an extracted snippet of a scientific document, referred to as the \textit{context}, which contains the target concept. In our experiments, the contexts are limited to 1-2 contiguous sentences. (ii) They must mention another concept, referred to as the \textit{reference} concept, which is mentioned in the extracted context and is related to the target concept by one of the four relations in $\{\texttt{is-a}, \texttt{is-like}, \texttt{part-of}, \texttt{used-for}\}$.  This relation must also be reflected in the extracted context. (iii) The description must contain an \textit{elaboration}, or a span of text that further explains the specified relation between the target and reference concepts. For example, a description cannot only say that ``SQuAD is like TriviaQA'' but it must also specify that they ``are both reading comprehension data sets.'' These elaborations must be supported by the associated extracted context.

The description criteria described above enabled us to build a system that produced many descriptions preferred by users, as we show in our experiments.  However, the DSG task is more general than our specific formulation, and experimenting with richer description formats in DSG is an important item of future work.

\section{Data set} \label{sec:dataset}

\begin{table}[t]
    \centering
    \begin{tabular}{p{0.35\linewidth} p{0.55\linewidth}}
    \toprule
    \footnotesize \textbf{Extracted context} & \footnotesize \textbf{Hand-authored descriptions} \\ \hline
    \multirow{3}{2.5cm}{\footnotesize word embedding is a \hlpink{word representation} that captures semantic and syntactic similarities between words. it has been widely utilized for a variety of tasks, such as  \hlpink{sentence classification} [42], \hlpink{relation classification} [41], and sentiment analysis [38], since the introduction of word2vec software. \cite{Shi2019RetrofittingCW}} & \footnotesize [\hlpink{sentence classification}, \hlpink{relation classification}] \hlyellow{is a} \hlgreen{task} \hlindigo{that word embedding has been utilized for since the introduction of word2vec software}. \\ \cdashline{2-2} 
    & \footnotesize \hlpink{sentence classification} \hlyellow{is like} [\hlgreen{relation classification}, \hlgreen{sentiment analysis}] \hlindigo{in that they are both tasks that word embedding has been used for since the introduction of word2vec software}. \\ \cdashline{2-2} 
    & \footnotesize \hlpink{relation classification} \hlyellow{is like} [\hlgreen{sentence classification}, \hlgreen{sentiment analysis}] \hlindigo{in that they are both tasks that word embedding has been used for since the introduction of word2vec software}. \\ \cdashline{2-2} 
    & \footnotesize \hlpink{word representation} has been \hlyellow{used for} [\hlgreen{sentence classification}, \hlgreen{relation classification}, \hlgreen{sentiment analysis}] since the introduction of word2vec software. \\ 
    \bottomrule
    \end{tabular}
    \caption{\textit{Sample entry from the ACCoRD corpus.} The ACCoRD annotation procedure uniquely allows each positively-labeled context to yield multiple concept descriptions for target ForeCite concept(s) (\hlpink{red}) present in an extracted context. Diversity among these concept descriptions is induced through multiple relation types (\hlyellow{yellow}) and distinct reference concepts (\hlgreen{green}), each with an elaboration that specifies the relationship between the target and reference concepts (\hlindigo{blue}).}
    \vspace*{-3mm}
    \label{tab:accord-corpus-example}
\end{table}

To support work on \taskacronymnospace, we compile and release the ACCoRD corpus.
The data set consists of 1,275 labeled contexts and 1,787 hand-authored concept descriptions, and induces diversity among these concept descriptions in two key ways. First, our data set allows for concept descriptions beyond the typical \texttt{is-a} relation.
Second, a single target concept is allowed to be described in terms of any number of other concepts in the source text.

\begin{figure*}[t!]
    \centering
    \includegraphics[width=\textwidth]{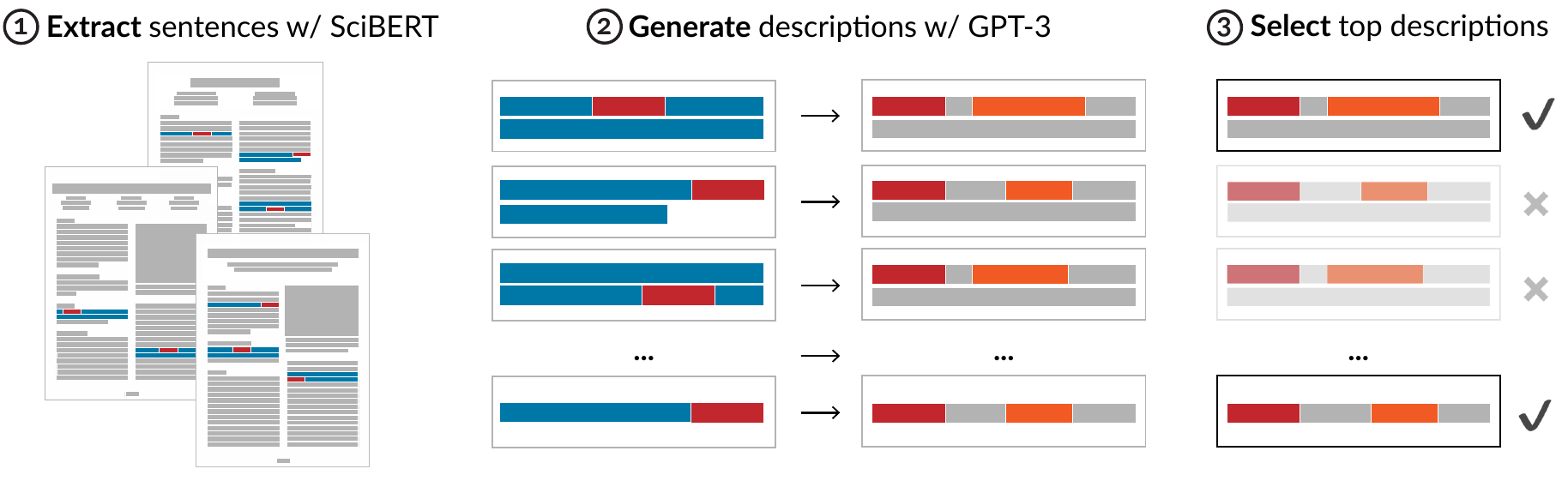}
    \caption{\textit{ACCoRD system implemetation.} Our system (1) \textbf{extracts} context sentences (blue) from scientific documents that describe a target scientific concept (red) in terms of another using SciBERT \cite{scibert} finetuned on the ACCoRD corpus, (2) \textbf{generates} succinct, self-contained, and distinct descriptions of the target's relationship to each reference concept (orange) from the extracted contexts using GPT-3 \cite{brown2020language} in the few-shot setting, and (3) \textbf{selects} a final description set involving multiple relation types and reference concepts.}
    \label{fig:system-schematic}
\end{figure*}

\subsection{Data set construction}

To construct the ACCoRD corpus, we consider the abstract, introduction, and related works sections of 698 CS papers from S2ORC~\cite{s2orc}. We automatically identify candidate contexts of 1-2 contiguous sentences with at least one significant CS concept by performing simple string matching against a high-precision set of CS concepts from ForeCite~\cite{forecite}. ForeCite assigns a score to candidate concept string using the citation graph, based on the intuition that when papers containing a string all tend to cite the same paper, that string is more likely to be a bona-fide scientific concept.  We filter to concepts with a ForeCite score $\geq 1.0$.

Annotators were instructed to assign a binary label to each candidate context indicating whether the context sentence(s) contained a description of the target ForeCite concept in terms of any other concept in the context. This component of our annotation task was found to have high inter-annotator agreement with a Cohen's $\kappa$ of $0.658$. 

For each extracted context that was assigned a positive label, annotators were instructed to author as many descriptions of the target ForeCite concept that follow criteria 1-3 above.\footnote{In a small fraction of cases the hand-labeled descriptions deviate from the criteria. First, in $<4.6\%$ of examples, we allowed annotators to specify a reference concept not explicitly mentioned in the extracted context. These were limited to obvious cases; e.g. ``neural network'' is a reference concept for target concept ``recurrent neural network.''  Second, $<0.4\%$ of examples do not contain an elaboration. The majority of these cases are of the \texttt{used-for} relation, where the reference concept and elaboration are a single entity (e.g. ``gav is used for \textit{query processing in stable environments}.'')}
These criteria uniquely allow each positively-labeled context to yield multiple concept descriptions if a target concept was described in terms of multiple other concepts in the source text, if a single concept pair can be described using multiple relations, or if the extraction contained multiple target concepts (see Table \ref{tab:accord-corpus-example}).

\section{System overview} \label{sec:system-overview}

The ACCoRD system has 3 pipeline stages: (1) \textbf{extract} sentences that describe one scientific concept in terms of another, (2) \textbf{generate} succinct, self-contained descriptions of the concepts' relationship, and (3) \textbf{select} the resulting descriptions to produce a final set of top descriptions for each concept (see Figure \ref{fig:system-schematic}).

\paragraph{Extraction}
We build a two-stage SciBERT-based~\cite{scibert} extractor to identify sentences that describe a target concept in terms of another concept. In the first stage, a binary classification model trained on the binary labels from the ACCoRD corpus extractions identifies reasonable candidates containing a description of the target concept in terms of another concept. In the second stage, a multilabel classification model trained on the relation types in ACCoRD predicts a relation type for a given extracted context, such that we can assign it to the appropriate prompt for the few-shot generation model described below. 
The inputs to both models have the target scientific concepts demarcated following \newcite{wu2019enriching}. We select optimal hyperparameters using cross-validation on the ACCoRD training set.  This two-stage approach was found to outperform using a single multi-label classifier in preliminary experiments. 

\paragraph{Generation} The positive-predicted outputs of our extraction model are used as input to our generation model, which produces succinct, self-contained summaries of the concept relationship described in the extracted text.
We use GPT-3's \code{davicini-instruct-beta} model \cite{brown2020language} in the few-shot setting.
We provide the model with a prompt that includes the instruction ``Describe the provided concept in terms of another concept in the text'' along with five hand-picked (extraction, ground truth concept description) example pairs from the ACCoRD corpus. We hand-select example pairs for each relation type, and for each query extraction, provide the examples of the relation type predicted by the multilabel classifier. 
This model maps each extraction to a single concept description.

Each generated description is then post-processed to heuristically identify the reference concept, using noun chunking and regular expressions based on our description templates.
We then apply additional heuristics to the descriptions as a first-pass filter for low-quality GPT-3 generations, e.g. removing descriptions that have any mention of unresolved references like ``our work,'' descriptions that erroneously contain a reference concept that is an author's name, and descriptions with more than one occurrence of the target concept (to prevent descriptions of the target concept in terms of itself).

\paragraph{Selection}
Having obtained a candidate set of descriptions for each concept, we attempt to identify a smaller, easily-consumable set that is most likely to be informative and globally descriptive of the target concept using a selection process. We first filter our descriptions to those that involve a reference concept from ForeCite~\cite{forecite}. 
For each (target, relation) pair, we then choose the most frequent $k$ reference concepts among the descriptions.
This ranking method reflects the intuition that the reference concepts that are more helpful for understanding a target concept will also be used relatively more frequently to describe that concept. 
We then select a top description for each (target, reference, relation) triple by selecting the one with the highest prediction score from our multilabel extraction model. 
Building sets of descriptions that contain distinct (target, reference, relation) triples allows us to maximize the number of potentially helpful connections that can be made between the target concept and concepts known to the user (see analysis in Section \ref{sec:diversity-analysis}).

\begin{table*}[t]
    \centering
    \renewcommand{\arraystretch}{.7}
    \begin{tabular}{p{0.45\linewidth} p{0.45\linewidth}}
    \toprule
    
      \textbf{\footnotesize Extraction} & \textbf{\footnotesize Generation} \\ \hline
      
       \footnotesize some such models, including \hlpink{variational autoencoders} (vaes) and generative adversarial networks (gans) [goodfellow et al., 2014, kingma and welling, 2013, rezende et al., 2014], learn an explicit low-dimensional manifold that approximates a natural signal class. \cite{Asim2020InvertibleGM} & \footnotesize \hlpink{variational autoencoder} \hlyellow{is like} \hlgreen{generative adversarial network} in that they are both models that learn an explicit low-dimensional manifold that approximates a natural signal class. \\ \hdashline
       
      \footnotesize we then compare the performance of rbm with other standard representation learning algorithms, including principal or independent component analysis (pca, ica), autoencoders (ae), \hlpink{variational autoencoders} (vae), and their sparse variants. we show that rbms, due to the stochastic mapping between data configurations and representations, better capture the underlying interactions in the system and are significantly more robust with respect to sample size than deterministic methods such as pca or ica. \cite{Tubiana2019LearningCR} & \footnotesize \hlpink{variational autoencoder} \hlyellow{is like} \hlgreen{autoencoders} in that they are both methods for representation learning and are more robust with respect to sample size than deterministic methods such as pca or ica. \\ \hdashline
       
      \footnotesize such controllable text generation models help in a wide range of application, e.g., dialogues systems [17] . backed by the recent success of deep generative models, such as \hlpink{variational autoencoders} (vaes) [10] , generative adversarial nets (gans) [5] , and autoregressive models [12] , existing models have made progress towards controllable text generation [15, 8, 9, 18]. \cite{Xu2019UnsupervisedCT} & \footnotesize \hlpink{variational autoencoder} \hlyellow{is like} \hlgreen{generative adversarial net} in that they are both deep generative models that have made progress towards controllable text generation. \\ \hdashline
       
      \footnotesize generative models in combination with neural networks, such as \hlpink{variational autoencoders} (vae), have gained tremendous popularity in learning complex distribution of training data by embedding them into a low-dimensional latent space. traditional vaes usually incorporates simple priors, e.g., a single gaussian, for regularizing latent variables. \cite{Zhao2019TruncatedGV} & 
      \footnotesize \hlpink{variational autoencoder} \hlyellow{is a} \hlgreen{generative model} that is used in combination with neural networks to learn complex distribution of training data by embedding them into a low-dimensional latent space. \\ \hdashline
      
      \footnotesize recently, deep generative models such as \hlpink{variational autoencoders} (vaes) (rezende et al., 2014) have become increasingly popular for modelling real-valued data, such as images. \cite{Zhao2019DeepGM} & \footnotesize \hlpink{variational autoencoder} \hlyellow{is a} \hlgreen{deep generative model} that is used for modelling real-valued data, such as images. \\ \hdashline
       
      \footnotesize latent variable models such as \hlpink{variational autoencoders} (kingma and welling, 2013) tend to better capture the global feature representation in data, but do not offer an exact density estimate as they maximize a lower bound of it. implicit generative models such as gans have recently become popular for their ability to synthesize realistic data (karras et al., 2018; engel et al., 2019). \cite{Das2019DimensionalityRF} & \footnotesize \hlpink{variational autoencoder} \hlyellow{is a} \hlgreen{latent variable model} that does not offer an exact density estimate. \\
      \bottomrule

    \end{tabular}
    \caption{\textit{Sample ACCoRD system output for target concept ``variational autoencoder.''} Extracted contexts are synthesized into generations that describe the target concept (\hlpink{red}) in terms of a reference concept (\hlgreen{green}) using a specified relation type (\hlyellow{yellow}).}
    \vspace*{-4mm}
    \label{tab:generative_results}
\end{table*}

\subsection{ACCoRD generated descriptions are diverse} \label{sec:diversity-analysis}

By identifying concept descriptions across the scientific literature, our system is able to capture a diversity of descriptions for a given target concept (see Table \ref{tab:generative_results}).
We measure this diversity for a set of 150 popular NLP concepts using two metrics: the number of candidate descriptions prior to the selection stage of our system and the number of unique reference concepts contained in those descriptions.\footnote{We report these statistics per relation type exhibited in the description. For brevity, we restrict this to the two most commonly observed relation types.}

\begin{figure}[t]
    \centering
    \includegraphics[width=.47\textwidth]{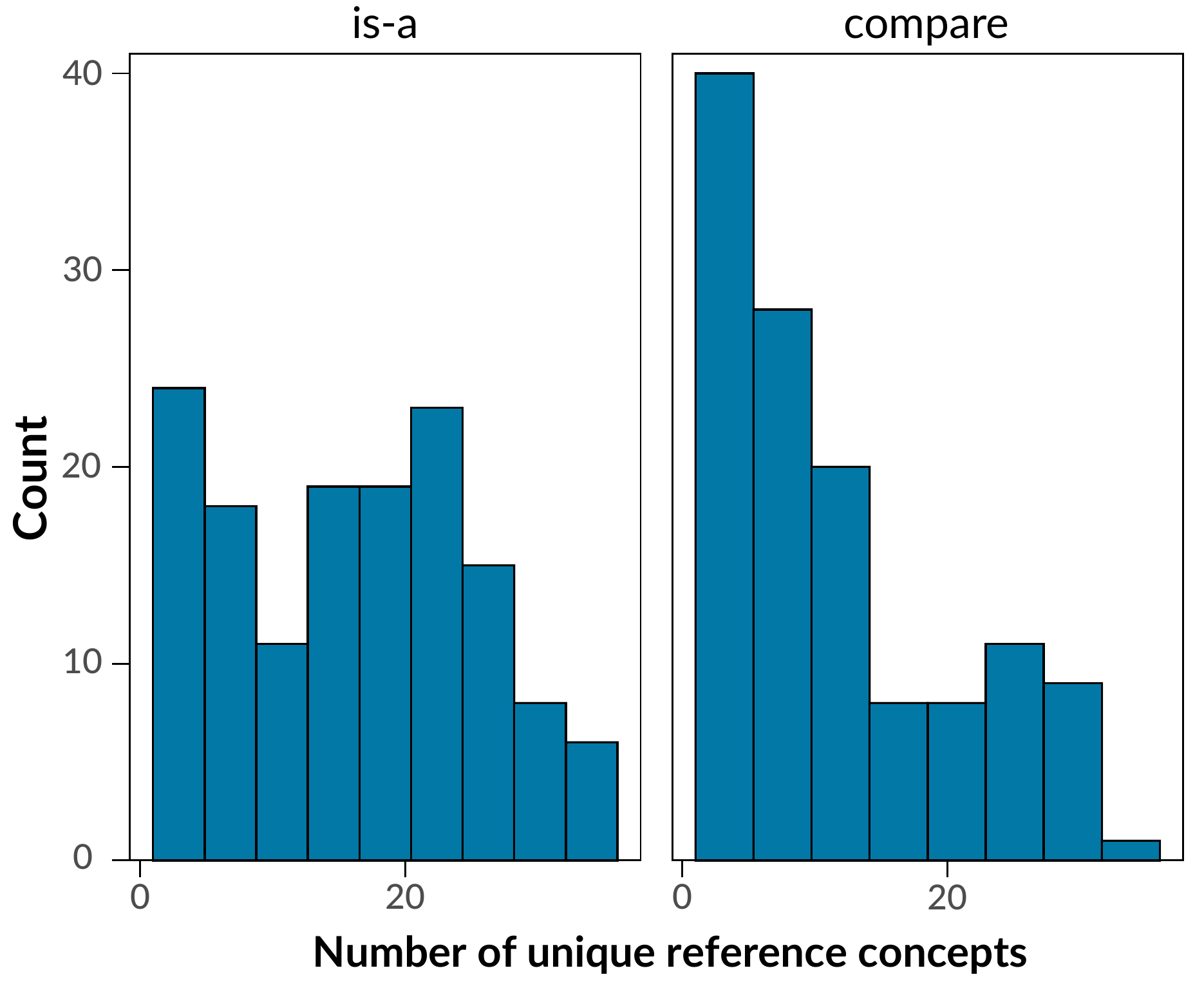}
    \caption{\textit{Distribution over number of unique reference concepts per target among 150 popular NLP concepts.} For each target concept, ACCoRD produces candidate descriptions involving a variety of reference concepts and relations.
    }
    \vspace*{-2mm}
    \label{fig:conceptb-hist}
\end{figure}

For descriptions involving the \texttt{compare} and \texttt{is-a} relations, we find an average of 153 and 373 candidate descriptions per target concept, respectively (see Appendix, Figure \ref{fig:description-counts}).
These candidate descriptions contain an average of 15 unique reference concepts per target concept for \texttt{is-a} descriptions and 11 for \texttt{compare} descriptions (see Figure \ref{fig:conceptb-hist}). 
Thus, our system captures a wealth of information from the scientific literature that is not retained by a ``single best'' approach.

\section{User study}
The experiments in the previous section show that our system produces meaningful diversity in generated descriptions.  In order to answer key questions about the utility of our methods, we perform a user study with the full end-to-end system.

\begin{itemize}
    \item \textbf{RQ1:} Which method of producing concept descriptions do users most prefer?
    \item \textbf{RQ2:} Does there exist a single ``best'' description per user? %
\end{itemize}

\subsection{Study description} \label{sec:study-description} 
In our study, participants were asked to imagine they were reading a section in a paper and came across a scientific concept they wanted to learn more about. Participants were asked their preferences for different \textit{sets} of descriptions, as well as their preferences for the individual descriptions.

\paragraph{Participants}
We recruited 22 participants with native or bilingual English proficiency whose areas of proficiency within computer science included natural language processing (NLP). Participants were recruited through Upwork. All participants had at least a bachelor's degree in computer science. 8 participants additionally had a Master's degree in computer science and 3 had obtained a PhD in computer science. 14 participants indicated having up to three years of experience in NLP, five had 4-6 years of experience, and three others indicated having more than 7 years of experience.

\paragraph{Design}
Our study consisted of two parts: the first was designed to understand users' preferences for \textit{sets} of descriptions, and the second aimed to understand their preferences for individual descriptions within the sets. For both parts, participants were asked to imagine they were reading a section in a paper and came across a scientific concept that they wanted to learn more or be reminded about.
A hypothetical interface was proposed that could provide multiple descriptions of these concepts while reading, in a pop-up box format. 

We selected a set of 20 popular NLP concepts with a ForeCite score greater than 1.0. For each concept, we obtained three sets of 6 descriptions each designed to test the various components of our approach:

\begin{itemize}
    \item \textbf{\setAname} the output of our complete system: generated descriptions that were selected according to our ranking and filtering methods. This set was comprised of the top 3 descriptions for each of the relation classes \texttt{compare} and \texttt{is-a}.
    \item \textbf{\setBname} the raw extractions corresponding to the generations in \setAname.
    \item \textbf{\setCname} the output of the generation component of our system, but without the final stratified selection step. Instead, the top 6 descriptions for this set were selected based on the prediction scores output by the extractive step of our system.
\end{itemize}

In the first part of the study, for each concept tested, participants were were first asked to indicate their level of expertise with the concept on a 5-point scale ranging from 1 = ``I do not know this concept'' to 5 = ``I know the concept and could explain it to someone else.''
They were then asked to read the three sets of descriptions for the given concept and select the description set that they deemed most helpful for the proposed task.
At the end of the section, participants were asked to describe how they evaluated/determined their preference for the sets of descriptions in a free-response style question. In particular, we asked them to articulate which features of the description sets were important in determining a preference, and how they evaluated the combination of the descriptions. 

In part two of the study, we were interested in understanding users' preferences at the individual description level. Participants were shown each of the 6 descriptions from our complete system's output (\setAname) and asked to indicate their level of preference for the description in one of the following three ways: ``I would want to see this description of the concept,'' ``No preference/opinion,'' ``I would not want to see this description of the concept.'' At the end of the section, participants were asked to respond to two free-response style questions explaining (1) why they preferred certain descriptions over others and the qualities that made them better or worse and (2) how their criteria may have differed when rating \textit{sets} of descriptions compared to \textit{individual} descriptions.

\subsection{Results} \label{sec:quant-findings}

\begin{figure}[t]
    \centering
    \includegraphics[width=.45\textwidth]{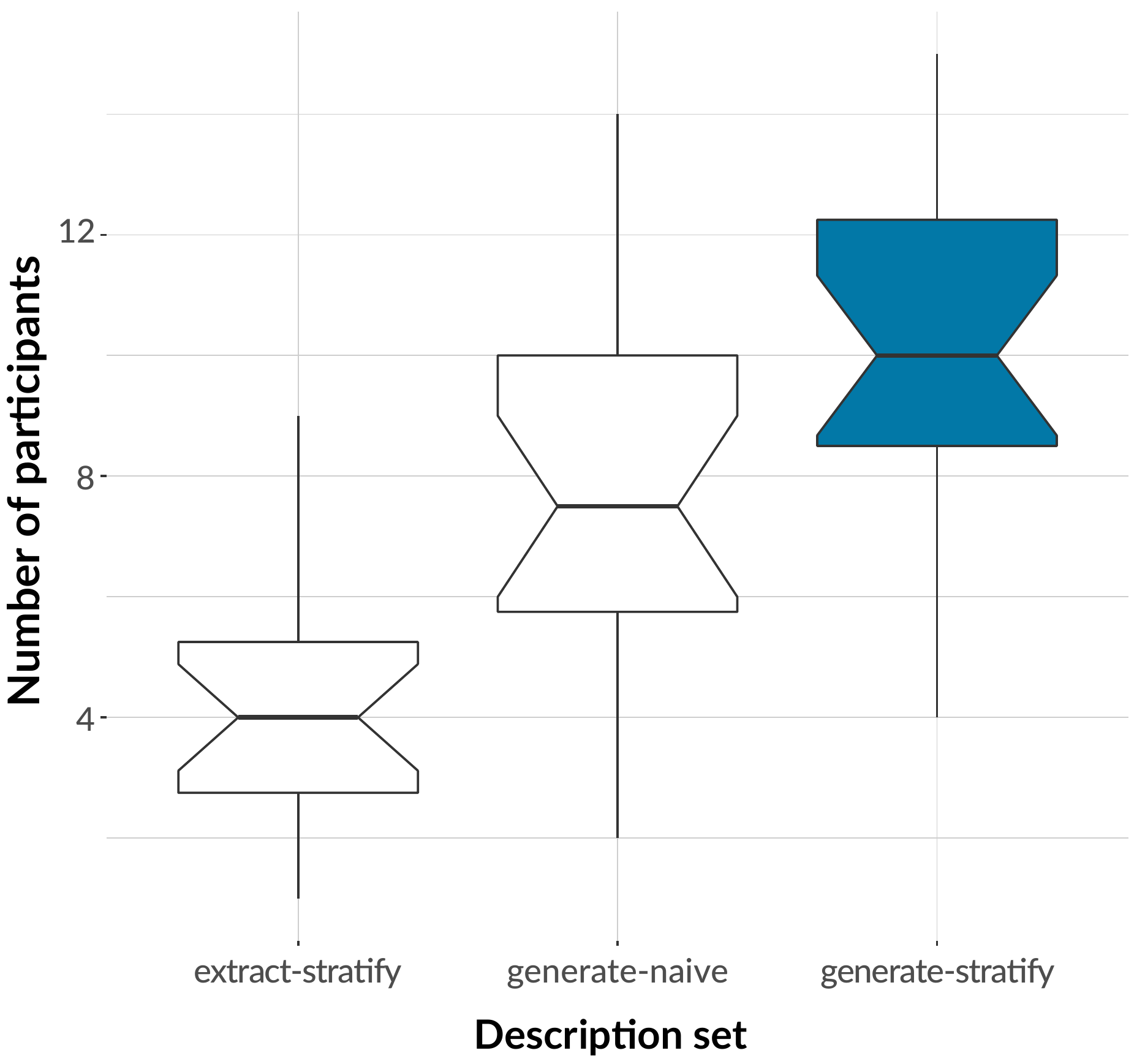}
    \caption{\textit{User preferences for description sets.} Participants strongly preferred descriptions sets that contained generated descriptions (\setAname, \setCname) over the set that contained extracted text snippets (\setBname). Participants' preference for the set produced by ACCoRD's more sophisticated description selection component was less pronounced, but still resulted in a higher minimum preference count than \setCname.}
    \vspace*{-2mm}
    \label{fig:set-preference-results}
\end{figure}

\paragraph{\textit{RQ1:} Users prefer our system's generated description over baselines}
The three description sets we tested were aimed at understanding users' preferences for the individual components of our system. In particular: (1) whether people preferred the final, summarized concept description over the raw extractions, and (2) whether our stratified selection method of filtering the concept descriptions was preferred.
 
As shown in Figure \ref{fig:set-preference-results}, aggregated over the responses for all 20 concepts in our study, participants strongly preferred both versions of the generated description sets, which received a median score of 7.5 for  \setCname\ (95\% CI = [5.9, 9.1]) and 10.0 for \setAname\ ([8.4, 11.6]) compared to the raw extractions from \setBname\ at 4.0 ([2.9, 5.1]). These results also show a preference for the description set obtained using ACCoRD's stratified selection method \setAname\ over \setCname.

\paragraph{\textit{RQ2:} There is no single ``best'' description per user}
\label{sec:rq2}
ACCoRD's approach is based on the hypothesis that users prefer a set of descriptions to a single ``best'' description per concept. Our findings support this idea: When presented with multiple individual descriptions from our end-to-end system, \setAname, participants on average preferred around 3 descriptions for a given target concept ($\mu=3.41$, 95\% CI = [3.03, 3.79]).

\subsection{Qualitative analysis}

An analysis of the free-text responses from study participants generally confirmed the results of our quantitative findings, while shedding more light on users' considerations in evaluating concept descriptions.

\paragraph{Users prefer concise descriptions} Participants most consistently articulated some preference for shorter, more concise, and more direct descriptions of the target concept ($n = 11$). This provides strong support for the generative component of our system, though, a number of users ($n = 5$) noted that the generations were not always accurate (see Section \ref{sec:future-work}). Additionally, though participants appreciated the simplicity and conciseness of the generated descriptions, many ($n = 6$) noted referencing the extracted text for additional context, confirming a design choice of our system to display each generation with its source text (see Figure \ref{fig:system-screenshots}).

\paragraph{Many users prefer analogical descriptions}
Our work expands the notion of a description in the context of description generation systems, to include analogy-like descriptions that are currently not captured by either scientific definition \cite{heddex} or relation extraction \cite{dygie} systems.
A number of participants ($n=9$) noted that descriptions that drew connections between other concepts in this fashion were helpful, in particular because they could ease learning and memorization of the concept (P18), reflected their own process when trying to synthesize new information (P19), and helped make sense of the many similar model architectures (P14). 

\section{Related Work}

\paragraph{Cognitive theories}
From children to experts, readers of all levels have been found to actively employ structured background knowledge in the process of comprehending a text ~\cite{Spiro1980, Bazerman1985PhysicistsRP}.
This background knowledge, often referred to as ``schemata,'' ``scripts,'' or ``frames'' in cognitive theories of knowledge representation, can be thought of as the data structure for storing concepts in human memory \cite{RumelhartOrtony1977}. These schemata contain the network of interrelations that hold for a given concept.
Cognitive theories of learning have asserted that effective ways of describing a new concept to someone take advantage of such schemata, by grounding new descriptions within the network of concepts they are already familiar with~\cite{nrc2018}. 
Systems that operate within the paradigm of providing a single ``best'' result for all users, as many definition generation systems do, limit the accessibility of technical knowledge to diverse audiences \cite{potentialPersonalization}.
These considerations motivate our system and novel task definition, which extends the conventional description generation setting to include multiple target descriptions for a single concept.

\paragraph{Data sets} 
While definition and relation extraction data sets address individual components of \taskacronymnospace, they each lack vital components that prevent them from being used as training data for the diverse, multi-relation view of concept descriptions that our task requires. 
Relevant to the scientific, computer science domain that we study is the W00~\cite{w00} data set, a corpus of 2,512 sentences from 234 workshop papers from the 2000 ACL Conference.
While this data set includes definitions that generalize beyond common lexico-syntactic patterns, they are ultimately restricted to definitions involving a typical \texttt{is-a} relationship between target and reference concept because of their traditional notion of a definition.
Data sets for evaluating scholarly relation extraction systems, like SciERC \cite{scierc}, do connect concept pairs using a variety of relation types. However, they do not include differentia between the concept pairs that elaborate on the concepts' relationships, making them unsuitable as data for the concept description setting we investigate.

\paragraph{Methods}
While previous work has investigated extracting and generating definitions of scientific concepts~\cite{scidef2,scidef3,scidef4,heddex}, they focus on producing a {\em single} canonical description for each concept. These methods also approach definition generation from a single-document perspective, which doesn't account for the multitude of ways a concept might be described outside of the context of that paper. In contrast, we aim to preserve multiple distinct descriptions that take advantage of the corpus-wide mentions of a given concept. In addition, we look to the related works sections of papers to identify descriptions that involve comparative relationships between concepts beyond the typical \texttt{is-a} relationship.

\section{Future work} \label{sec:future-work}

\subsection{Improving generation quality}
Our results show that a generation component that produces succinct, direct descriptions of a target concept is helpful for a user-friendly system for \taskacronymnospace. However, our qualitative feedback suggests that this is also an important area for future work, as poor generation quality was often cited as the reason users preferred the set with only extracted descriptions. 

An analysis of a sample ($n=100$) of ACCoRD system outputs revealed particular areas for improvement. $48\%$ of the sample contained at least type of error.
The majority of the error came from the generation component of our system and included generations that were an inaccurate synthesis of extracted context; generations that were technically correct, but unhelpful because they lacked important details; incoherent generations; instances where the second sentence of extraction was appended verbatim, often resulting in one of the other error types; and generations where irrelevant details were appended  (see Table \ref{tab:error_analysis_generation} for proportions and an example of each error type). 
In a minority of cases, system errors were due to issues at the extractive stage. These error types included poor delimitation of the target ForeCite concept within candidate extractions that resulted in an inaccurate generation; low quality extracted text that made it difficult to produce a high quality generation; and difficulties in delimiting sentences in scientific text using available methods (see Table \ref{tab:error_analysis_extraction}). While these errors are of varying degrees of severity, further work on methods that address these issues at scale will be important to expanding the proof-of-concept system we propose here. 

\subsection{Controllable generation}
Beyond resolving errors in generation, future work might investigate methods for controllable generation that are better tailored to user needs.  For example, in our user study free-text responses, participants suggested that users may require different kinds of descriptions based on the type of concept being described. In particular, two participants (P14, P21) noted that they preferred set \setCname, which contained more "canonical" descriptions, for simple, standalone data set concepts that could be explained straightforwardly. On the other hand, for more complex method and system-based concepts, like RoBERTa, GPT, and LSTM, users expressed preference for descriptions that made comparisons to other concepts (as produced by our complete system).

\subsection{Potential for personalization}

While we showed in Section \ref{sec:rq2} that participants often had multiple preferred descriptions per concept, a question remains---Are these preferences similar or different \emph{across} users?
To investigate, we compute the Fleiss' $\kappa$ score measuring agreement in participant preference votes across the six available descriptions for each concept, and find this to be low on average across concepts ($\mu=0.06$, 95\% CI = [0.04, 0.09]).  Likewise, only a minority of users ($\mu=0.34$, 95\% CI = [0.31, 0.36]) listed the top-voted description among their preferred ones, on average.  The high variation in preferences across participants suggests potential for personalization in the DSG task. In this section, we investigate avenues that future researchers might explore in order to operationalize personalization within description generation systems. 

\begin{figure}[h]
    \centering
    \includegraphics[width=.45\textwidth]{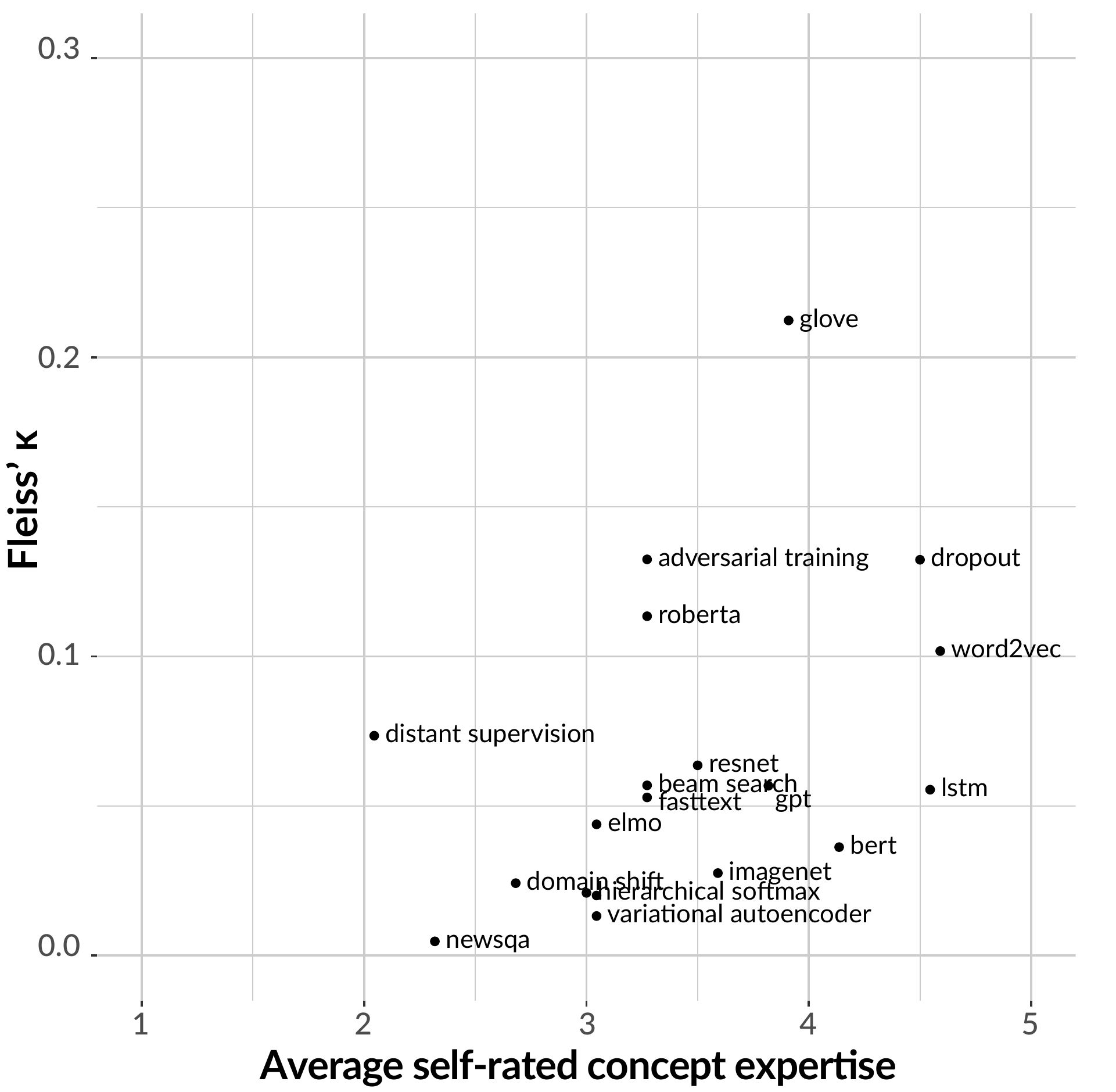}
    \caption{\textit{User agreement (Fleiss' $\kappa$) in description preferences for each concept versus average concept expertise level.} We find low agreement in preferences overall, with the lowest agreement scores for concepts for which participants also indicated low expertise.}
    \vspace*{-2mm}
    \label{fig:expertise-kappa}
\end{figure}

We consider how level of expertise in a concept might affect a user's preferences over descriptions. In Figure~\ref{fig:expertise-kappa}, we plot the Fleiss' $\kappa$ scores for user preferences over descriptions against the average level of self-reported expertise of the concepts.  While we do observe low agreement in preferences overall, interestingly the lowest agreement scores are found for concepts for which participants mostly self-rated as having low expertise.  Fitting a linear model to the data, we find the estimated slope coefficient is significantly greater than zero ($b = 0.03628, p < 0.05$).\footnote{We further investigated this by, for each concept, segmenting participants into high and low expertise groups (i.e. above or below the global median expertise of 4.0) and computing average agreement within those groups.  The difference in mean agreement between high ($\mu=0.08$) and low ($\mu=0.05$) expertise groups was not statistically significant.} In terms of qualitative evidence, we found in interviews that user study participants might even disagree on the manner in which expertise should affect ideal description output.  For example, P9 suggested that analogical descriptions would be most useful for concepts with low familiarity. In contrast, P14 stated that they would be more helpful for concepts with some level of expertise.

\section{Conclusion}

We have presented ACCoRD, an end-to-end system for the novel task of Description Set Generation (DSG): generating multiple distinct descriptions of a single target concept. 
In experiments, our methods were preferred over baseline approaches and produce a diversity of generated concept descriptions.
We also release the ACCoRD corpus to facilitate development of future systems for DSG. We hope that such systems will help increase the accessibility of scientific literature for people with diverse background knowledge.

\section*{Acknowledgements}
This project was supported in part by NSF Convergence Accelerator Award 2132318, NSF RAPID grant 2040196, ONR grant {N00014-18-1-2193}, and the Allen Institute for Artificial Intelligence (AI2). We thank Elena Glassman for providing valuable feedback on our system demonstration. 

\bibliography{anthology,custom}
\bibliographystyle{acl_natbib}

\newpage
\onecolumn
\section*{Appendix}

\subsection*{ACCoRD addresses a meaningfully novel task}
To verify that the ACCoRD corpus addresses a novel task that is not well-captured by existing resources, we compare our system's results on ACCoRD to those of existing state-of-the-art scientific definition and relation extraction systems. For our definition extraction baseline, we test HEDDEx~\cite{heddex} trained on W00~\cite{w00}, a similarly-sized corpus of definition sentences from workshop papers from the 2000 ACL Conference. Since HEDDEx was originally only intended to produce a single canonical definition of scientific terms and symbols at the sentence-level, we also evaluate its performance on the subset of ACCoRD that was marked as containing an “is-a” relationship between the reference and target concept, to more faithfully evaluate its potential. For our relation extraction baseline, we test DyGIE++~\cite{dygie} trained on SciERC~\cite{scierc}, a scientific relation extraction data set. Table \ref{tab:extractive_results} shows these results for the union of the 1- and 2-sentence source text settings in ACCoRD, as our qualitative conclusions remained unchanged across these settings. Our model trained on ACCoRD outperforms models that target related tasks, even when they beat a baseline that always assigns positive labels, suggesting that our data set addresses an importantly different task.

\begin{table}[h]
    \centering
    \begin{tabular}{p{0.2\linewidth} p{0.2\linewidth} p{0.06\linewidth}}
    \toprule
       Model & Train set & F1 \\ \hline
       HEDDEx & W00 & 0.329\\
       HEDDEx\textsubscript{\emph is-a} & W00 & 0.449 \\
       DyGIE++ & SciERC & 0.532\\
       \textbf{SciBERT} & \textbf{ACCoRD} & \textbf{0.624} \\ \hline
       \multicolumn{2}{c}{Positive baseline} & 0.484\\
       \bottomrule
    \end{tabular}
    \caption{Results for our extractive model and relevant baselines on the ACCoRD test set $(n=674)$. Our model trained on ACCoRD outperforms models that target related tasks, even when they beat a baseline that always assigns positive labels, suggesting that our data set addresses an importantly different task.}
    \vspace*{-3mm}
    \label{tab:extractive_results}
\end{table}

\begin{table*}[t]
    \centering
    \renewcommand{\arraystretch}{.7}
    \begin{tabular}{p{0.1\linewidth} p{0.1\linewidth} p{0.35\linewidth} p{0.35\linewidth}}
    \toprule
    
      \textbf{\footnotesize Error type} & \textbf{\footnotesize Percentage} & \textbf{\footnotesize Sample extraction} & \textbf{\footnotesize Sample generation} \\ \hline
    
       \footnotesize Generation is an inaccurate synthesis of extraction &  \footnotesize 24\% & \footnotesize in the modern context, mt systems can be categorized into statistical machine translation (smt) and \hlpink{neural machine translation} (nmt). smt has had its share in making mt very popular among the masses. \cite{Mahata2019JUMTAW} & \footnotesize \hlpink{neural machine translation} is like statistical machine translation in that they are both types of mt systems and have had their share in making mt very popular among the masses. \\ \\
       
       & & \footnotesize in the nlp community, \hlpink{beam search} and importance sampling are common. unfortunately, \hlpink{beam search} considers only the approximate-top-k taggings from an exponential set (wiseman and rush, 2016) , and importance sampling requires the construction of a good proposal distribution (dyer et al., 2016) . \cite{Lin2018NeuralPS} & \footnotesize \hlpink{beam search} is like importance sampling in that they are both methods that consider only the approximate-top-k taggings from an exponential set and require the construction of a good proposal distribution. \\ \hdashline
       
       \footnotesize Generation is technically correct, but unhelpful because it lacks important details &  \footnotesize 15\% & \footnotesize however, these approaches lack geometric priors and contain no mechanism for reasoning about the likelihood of alternative trajectories -a crucial skill for the task, e.g., 'would this look more like the goal if i was on the other side of the room?'. due to this limitation, many previous works have resorted to performing inefficient first-person search through the environment using search algorithms such as \hlpink{beam search} [4, 6] . \cite{Anderson2019ChasingGI} & \footnotesize \hlpink{beam search} is a search algorithm that is used for first-person search through the environment. \\ \hdashline
       
      \footnotesize Generation is incoherent &  \footnotesize 8\% & \footnotesize in recent years, generative models like \hlpink{vae} and gan [20, 11] and their extensions have become popular and have been applied to various scenarios due to their impressive performance. since the arterial and venous phases form different data distribution, to build a relation between them refers to the domain adaptation using generative models, where pix2pix [19] , cyclegan [50] and unit [28] etc. are typical models in this field. \cite{Zheng2018PhaseCN} & \footnotesize \hlpink{vae} is like gan in that they are both generative models that have been applied to various scenarios due to their impressive performance and since the arterial and venous phases form different data distribution, to build a relation between them refers to the domain adaptation using generative models.\\ \\
      
      && \footnotesize the trained model can be fine-tuned for downstream nlp tasks such as \hlpink{question answering} and language inference without substantial modification. bert outperforms previous state-of-the-art models in the eleven nlp tasks in the glue benchmark (wang et al., 2018) by a significant margin. \cite{Jawahar2019WhatDB} & \footnotesize \hlpink{question answering} is like language inference in that they are both nlp tasks that bert outperforms previous state-of-the-art models in the eleven nlp tasks in the glue benchmark (wang et al., 2018) by a significant margin. \\ \hdashline
      
    \footnotesize Second sentence of extraction is appended verbatim, resulting in a difficult to read, incoherent, or inaccurate generation &  \footnotesize 6\% & \footnotesize recent statistical machine translation (smt) systems employ a linear combination of several model components, such as translation models, \hlpink{language models}, and reordering models. translation is performed by selecting the most-likely translation, which is the candidate translation with the highest score based on the different model components. \cite{Suzuki2011DistributedME} & \footnotesize \hlpink{language model} is like translation model in that they are both model components that are used in recent statistical machine translation (smt) systems and translation is performed by selecting the most-likely translation, which is the candidate translation with the highest score based on the different model components. \\ 
      
    \bottomrule
    \end{tabular}
    \caption{\textit{Error analysis of ACCoRD system output: generation stage.} Common generation error types for a sample ($n=100$) of ACCoRD system outputs include generations that are an inaccurate synthesis of source text; generations that are technically correct but unhelpful because they lack important details; incoherent generations; and instances where the second sentence of extraction is appended verbatim, often resulting in a difficult to read, incoherent, or inaccurate generation. A final error type not shown was, appending irrelevant information, which occurred in 5\% of the sample. Multiple error types can be present in the same generation. }
    \vspace*{-4mm}
    \label{tab:error_analysis_generation}
\end{table*}

\begin{table*}[t]
    \centering
    \renewcommand{\arraystretch}{.7}
    \begin{tabular}{p{0.1\linewidth} p{0.1\linewidth} p{0.35\linewidth} p{0.35\linewidth}}
    \toprule
    
      \textbf{\footnotesize Error type} & \textbf{\footnotesize Percentage} & \textbf{\footnotesize Sample extraction} & \textbf{\footnotesize Sample generation} \\ \hline
      
     \footnotesize Poor target concept delimitation resulting in inaccurate generation & \footnotesize 6\% & \footnotesize in order to ensure accurate word alignments, most systems employ a post-hoc symmetrization step to combine directional word aligners, such as \hlpink{ibm model 4} (brown et al., 1993) or hidden markov model (hmm) based aligners (vogel et al., 1996) . several authors have proposed bidirectional models that incorporate this step directly, but decoding under many bidirectional models is np-hard and finding exact solutions has proven difficult. \cite{Chang2014ACV} & \footnotesize \hlpink{ibm model} is like hidden markov model in that they are both directional word aligners and have been used to create post-hoc symmetrization steps. \\ \hdashline
             
      \footnotesize Extracted text is missing details or contains unresolved references that make it difficult to produce a high quality generation &  \footnotesize 6\% & \footnotesize most of them generalize widespread existing unilayer measures such as degree [29, 17, 1] , betweenness [28, 34, 11] , closeness [29, 35] , eigenvector [17, 33, 1] , \hlpink{pagerank} [31, 20, 13] or hits [26] . these generalizations rely on the adaptation of unilayer concepts to the multilayer case. \cite{ReiffersMasson2017OpinionbasedCI} & \footnotesize \hlpink{pagerank} is like degree, betweenness, closeness, eigenvector and hits in that they are all unilayer measures that generalize to the multilayer case. \\ \\
      
      && \footnotesize [22] . their system achieved good results on various labeling tasks, including \hlpink{ner} and pos, without depending on any manual feature engineering. \cite{Ali2019BoostingAN} & \footnotesize \hlpink{ner} is like pos in that they are both labeling tasks that the system achieved good results on without depending on any manual feature engineering. \\ \hdashline
      
      \footnotesize Poor sentencization of scientific text & \footnotesize 4\% & \footnotesize various approaches have been proposed for sequence-tosequence problems, such as speech recognition [15, 16] , \hlpink{neural machine translation} [17, 18] , and dialogue modeling [19, 20] . among various deep-learning-based methods, cho et al. \cite{Lu2017OrderPreservingAS} & \footnotesize \hlpink{neural machine translation} is like speech recognition in that they are both approaches for sequence-tosequence problems and among various deep-learning-based methods, cho et al. \\
      
    \bottomrule
    \end{tabular}
    \caption{\textit{Error analysis of ACCoRD system output: extraction stage.} Common extraction error types for a sample ($n=100$) of ACCoRD system outputs include poor delimitation of the target ForeCite concept within candidate extractions, resulting in an inaccurate generation; low quality extracted text that results in low quality generation; and poor sentence delimitation of scientific text. Multiple error types can be present in the same generation.}
    \vspace*{-4mm}
    \label{tab:error_analysis_extraction}
\end{table*}

\begin{figure*}
\centering
\begin{subfigure}{\textwidth}
  \centering
  \includegraphics[width=\textwidth]{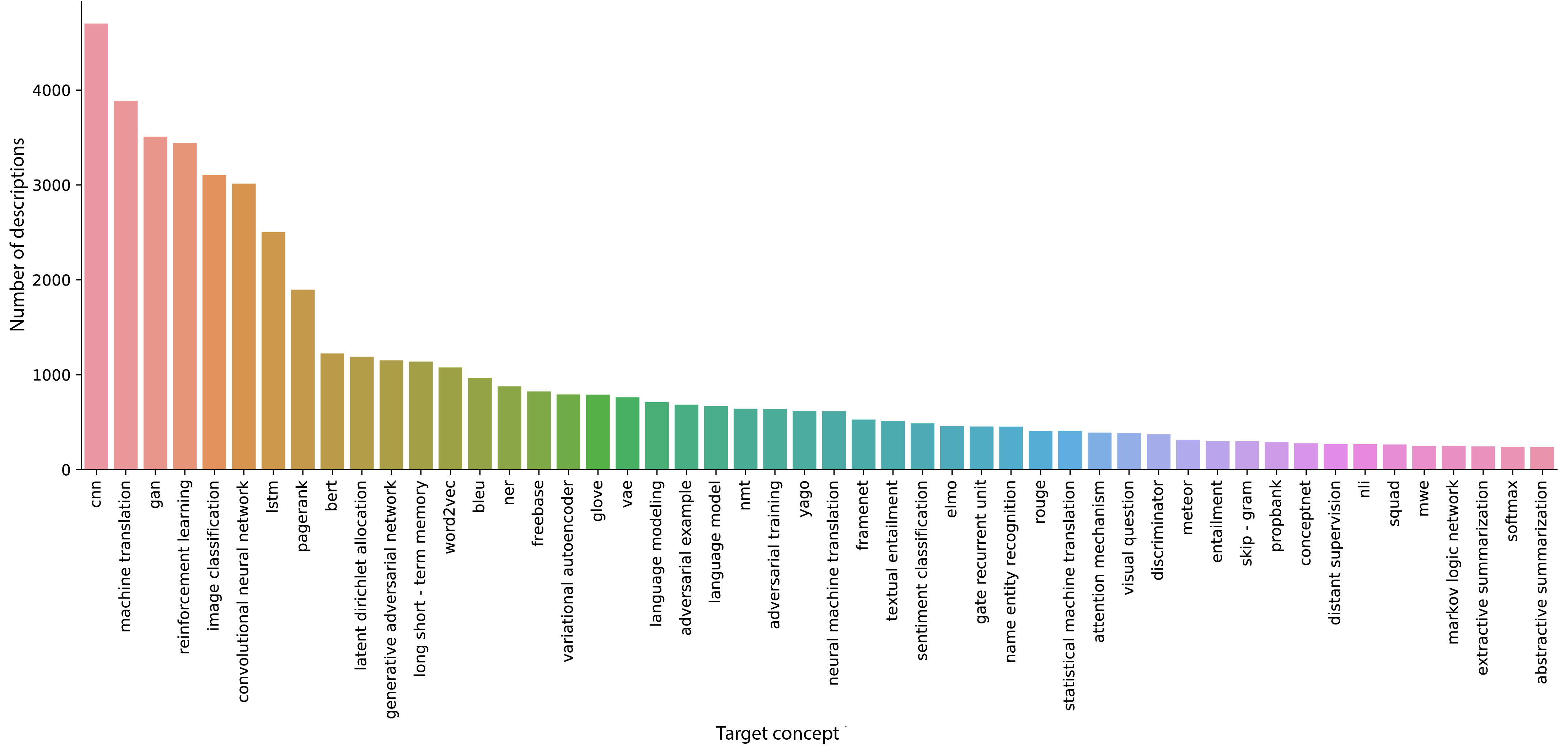}
  \caption{\texttt{compare} relation}
  \label{fig:sub1}
\end{subfigure}%
\\
\begin{subfigure}{\textwidth}
  \centering
  \includegraphics[width=\textwidth]{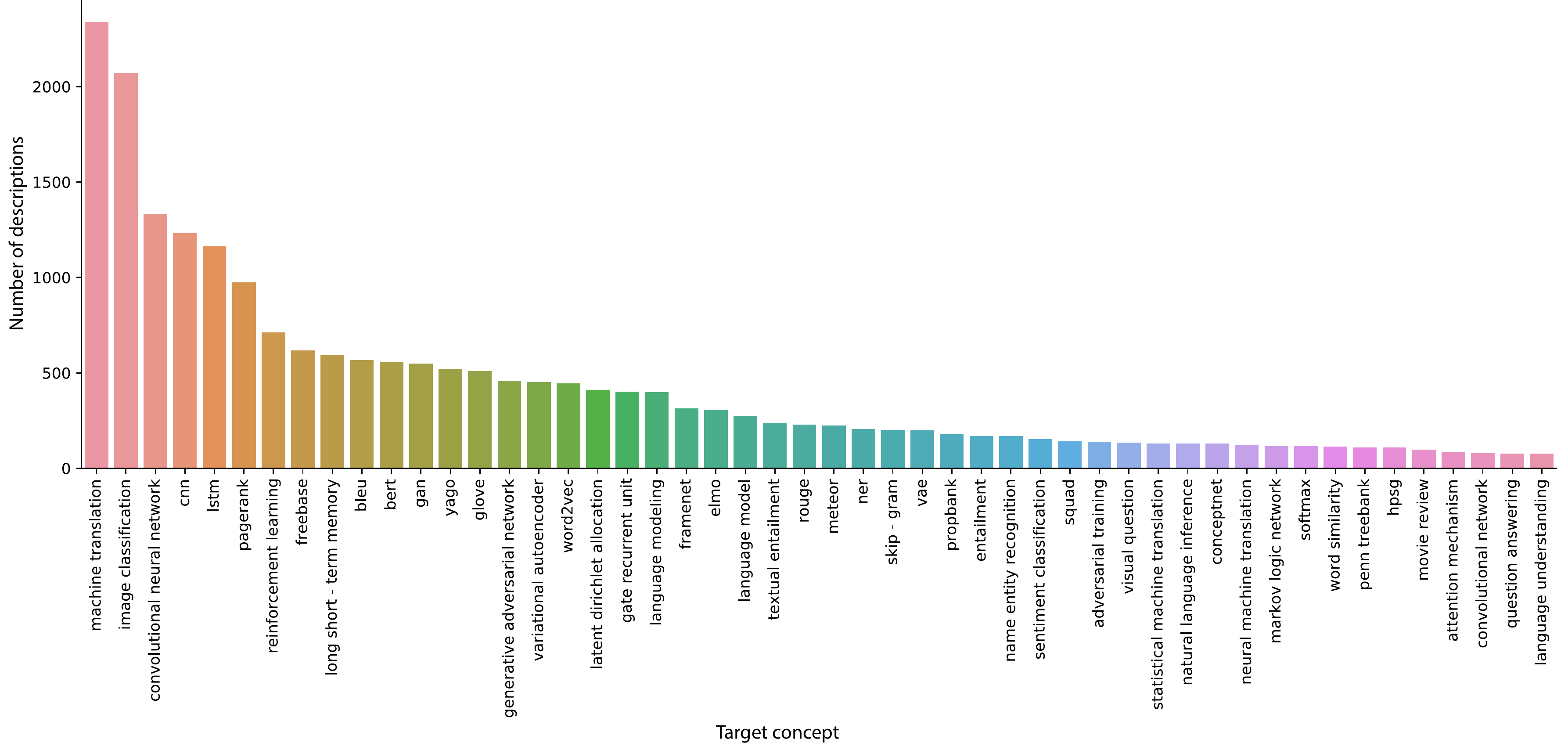}
  \caption{\texttt{is-a} relation}
  \label{fig:sub2}
\end{subfigure}
\caption{Number of candidate descriptions for 50 target NLP concepts, for each relation type present in our system demo, prior to the selection stage of ACCoRD. Our system identifies, extracts, and generates approximately twice as many candidate descriptions of the \texttt{compare} relation than of the \texttt{is-a} relation.}
\label{fig:description-counts}
\end{figure*}

\end{document}